\documentclass[conference]{IEEEtran}
\IEEEoverridecommandlockouts
\usepackage{cite}
\usepackage{amsmath,amssymb,amsfonts}
\usepackage{algorithmic}
\usepackage{graphicx}
\usepackage{textcomp}
\usepackage{subcaption}
\usepackage{url}
\usepackage{float}

\usepackage{xcolor}
\def\BibTeX{{\rm B\kern-.05em{\sc i\kern-.025em b}\kern-.08em
    T\kern-.1667em\lower.7ex\hbox{E}\kern-.125emX}}
\begin{document}

\title{EFFGAN: Ensembles of fine-tuned federated GANs
}

\author{\IEEEauthorblockN{1\textsuperscript{st} Ebba Ekblom}
\IEEEauthorblockA{\textit{RISE Research Institutes of Sweden} \\ 
\texttt{ebba.ekblom@ri.se}}
\and
\IEEEauthorblockN{2\textsuperscript{nd} Edvin Listo Zec}
\IEEEauthorblockA{\textit{RISE Research Institutes of Sweden}\\ 
\texttt{edvin.listo.zec@ri.se}}
\and
\IEEEauthorblockN{3\textsuperscript{rd} Olof Mogren}
\IEEEauthorblockA{\textit{RISE Research Institutes of Sweden}\\ 
\texttt{olof.mogren@ri.se}}
}

\maketitle
\begin{figure*}[t]
    \centering
    \includegraphics[width=\textwidth, height=15em]{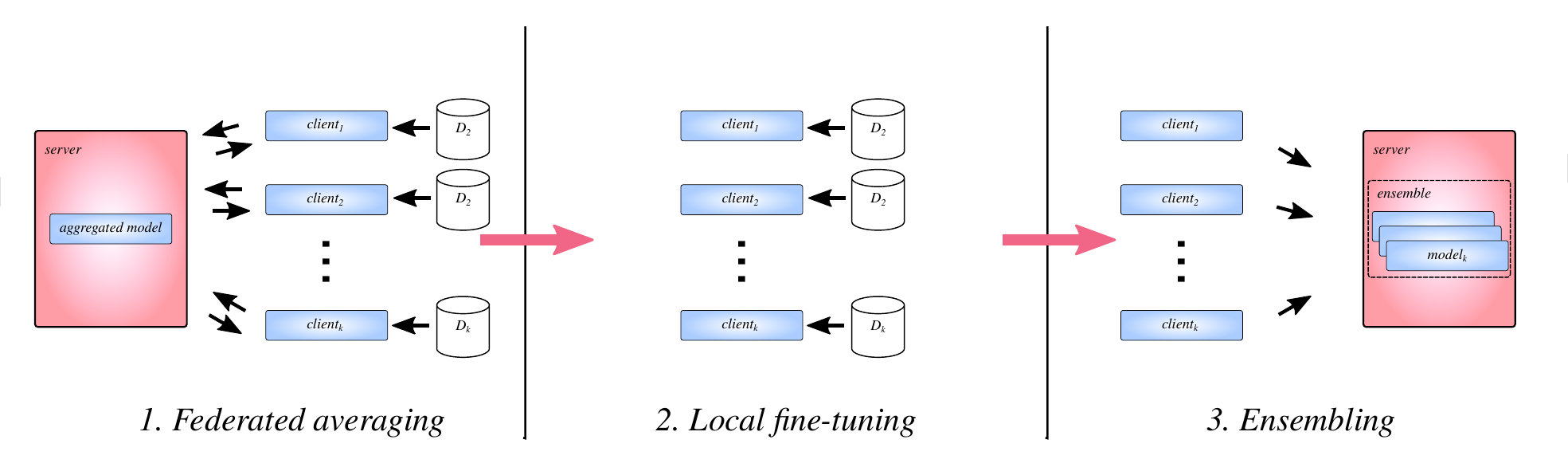}
    \caption{Schematic of the EFFGAN training protocol. The solution leverages the combination of federated learning with local fine-tuning and ensembling to produce a generative model trained on decentralized private data.}
  \label{fig:teaser}
\end{figure*}
\begin{abstract}
Decentralized machine learning tackles the problem of learning useful models when data is distributed among several clients. The most prevalent decentralized setting today is federated learning (FL), where a central server orchestrates the learning among clients. In this work, we contribute to the relatively understudied sub-field of generative modelling in the FL framework.
  
We study the task of how to train generative adversarial networks (GANs) when training data is heterogeneously distributed (non-iid) over clients and cannot be shared. Our objective is to train a generator that is able to sample from the collective data distribution centrally, while the client data never leaves the clients and user privacy is respected. We show using standard benchmark image datasets that existing approaches fail in this setting, experiencing so-called client drift when the local number of epochs becomes to large and local parameters drift too far away in parameter space. To tackle this challenge, we propose a novel approach named \textit{EFFGAN: Ensembles of fine-tuned federated GANs}. Being an ensemble of local expert generators, EFFGAN is able to learn the data distribution over all clients and mitigate client drift. It is able to train with a large number of local epochs, making it more communication efficient than previous works.
\end{abstract}


\section{Introduction}
Phones and tablets have increasingly been getting more powerful in the last decades. Today, they carry strong sensors such as cameras, microphones and GPS. People around the world frequently carry these devices wherever they go, and as such, these devices contain a lot of data. This data, due to the nature of how it is collected, is very private and cannot be shared or collected centrally. Meanwhile, it still could be leveraged and greatly improve user applications. It is in this setting that federated learning \cite{mcmahan2017communication} has become a paradigm shift for distributed machine learning. Federated learning is a framework for machine learning using decentralized data, allowing the training data to remain at the clients. Yet the data from many clients can be used in training the resulting machine learning model. Since federated averaging (\textsc{FedAvg}) was introduced in \cite{mcmahan2017communication}, the amount of research on it has increased at a fast pace. Problems such as heterogeneous client data, heterogeneous systems, privacy, fairness and many more are active research areas. At the same time, most of the current work focuses on mainly on \textit{supervised} tasks, such as classification. Much less has been done in the realm of self-supervised and unsupervised learning \cite{kairouz2019advances}.

In this work, we study unsupervised learning in the form of generative adversarial networks (GANs) \cite{goodfellow2014generative} in a distributed setting. We investigate the application of federated learning to GANs, and conclude that while existing federated learning strategies for them work well when training data is identically and independently distributed (iid), they fail to produce robust models when training data is heterogeneous (non-iid).

GANs are unsupervised generative models where two neural networks, a generator and a discriminator, compete in a two-player minimax game. They were proposed as a powerful approach to modelling complex and high-dimensional data distributions. Since the inception of GANs, a number of issues have been identified and addressed leading to more recent versions which suffer less from problems such as mode collapse, convergence failure, and instability during training. However, the decentralized setting of GANs, where data is distributed among a number of clients, has received much less attention in the literature. A decentralized GAN would allow for more powerful generative models trained on data from devices such as smartphones and tablets while at the same time respecting the privacy of users' data, as it does not need to be collected centrally. Meanwhile, training these models decentralized brings forth a lot of challenges. GANs are usually data-hungry, requiring a lot of data to perform well. Even when data is sufficient, instability issues are frequent even in central training. Further, data collected in the real world using devices such as phones and tablets is naturally non-iid. Label distribution skew and covariate shift are common distributional shifts that need to be addressed. 

Therefore, in this paper, we investigate the question of how to learn a data distribution over decentralized data in a non-iid setting. We provide an experimental evaluation where data from standard benchmark datasets (MNIST and FashionMNIST) are heterogeneously distributed over a set of clients, where certain classes of objects occur only on subsets of the clients. Finally, we propose a novel solution for distributed training of generative adversarial networks in the setting of heterogeneous data. In this setting, training local models on each client results in models that overfit to the local datasets, and a model trained using vanilla federated learning (FedGAN \cite{rasouli2020fedgan}) on the same data results in an aggregate model which fails to produce convincing samples due to a too large number of local epochs, i.e. \textit{client drift}. Our proposed solution, \textit{EFFGAN: Ensembles of fine-tuned federated GANs} mitigates both issues. By building an ensemble of fine-tuned generators it provides results that outperform previous works. We find that EFFGAN robustly handles heterogeneous data settings and requires substantially less communication compared to the baseline.

In summary, our main contributions are
\begin{enumerate}
    \item We scale up previous work, and increase the number of clients from 5 to 100.
    \item We show that distributed GAN training in non-iid data settings suffers from client drift, i.e. the parameters of the locally trained models drift apart, resulting in a bad aggregated global model.
    \item We show that \textsc{FedAvg} possesses good transfer learning capabilities in the GAN setting, even though it suffers from client drift.
    \item Using these insights, we propose \textit{EFFGAN}, an ensemble that consists of fine-tuned generators trained from \textsc{FedAvg} which mitigates client drift and achieves better FID score as compared to previous works in non-iid data settings.
\end{enumerate}

Our goal is to learn a model that can be used centrally to generate as real data as possible. Thus, the final model will be used at the server and not at the clients (which is typically the goal in FL). This is useful for organizations such as hospitals and financial institutions, where acquiring data can be time-consuming and expensive, and where sharing data is hard or prohibited. A generative model which can be sampled from can thus facilitate many such problems.


\section{Related work}
The work in this paper is related to federated learning, generative adversarial networks and ensemble models. Federated learning was proposed for communication efficient distributed learning \cite{mcmahan2017communication}. The algorithm, called federated averaging, runs parallel stochastic gradient descent for a number of communication rounds on a subset of all the clients. After each round, the updated client models are sent to a central server where they are averaged and communicated back. Most work on federated learning has been focused on the supervised setting where clients have labeled data \cite{kairouz2019advances}. 

\textbf{Federated GANs.} Up until recently, semi-supervised and unsupervised methods have been lacking in the federated learning literature. We contribute to this area of research in this work by studying GANs \cite{goodfellow2014generative} in a distributed setup, which are trained unsupervised without any need of labels. 

For iid data settings, MD-GAN \cite{hardy2019md} is a solution based on a single generator at the server with distributed discriminators. The generator communicates generated data to all participating clients, which is a communication-intensive operation and not scalable to many clients.

Non-iid data is tackled in \cite{yonetani2019decentralized}, where the discriminators are distributed and the server keeps a central generator, just as for MD-GAN. Here, the generator is updated using the most forgiving discriminator (i.e. the largest loss). Augenstein et al. \cite{augenstein2019generative} propose a similar federated GAN, with only one central generator and several distributed discriminators. The difference is that the discriminator update is based on selecting a subset of clients. Both these works suffer from large communication overhead, as the central generator needs to send data to each participating client each communication round.

FedGAN \cite{rasouli2020fedgan} is a GAN trained across distributed clients with non-iid data. The generator in FedGAN does not provide the same bottleneck effect, as each client has its own local generator and discriminator which are updated with federated averaging through the central server. FeGAN \cite{guerraoui2020fegan} is a similar solution, which also uses federated averaging on the generator and the discriminator. Here, the authors however aggregate models using Kullback-Leibler weighting, which uses a softmax on the KL divergence score of each client. Each model $w_k$ from client $k$ is assigned a weight 
\begin{equation}
\alpha_k = \frac{e^{-s_k}}{\sum_i e^{-s_i}},
\end{equation}
where $s_k$ is the pre-computed KL score of client $k$. FeGAN further utilizes balanced sampling every communication round, i.e. sampling clients with balanced datasets and more samples. However, this breaks some privacy concerns as clients need to disclose information about their data distributions to the central server.

Gossip GAN \cite{hardy2018gossiping} is a completely decentralized approach, where there is no central server available and where the clients communicate in a peer-to-peer network.

\textbf{Ensembles of deep nets.} Ensembles of deep neural networks have been studied in a non-federated setup. Deep ensembles were introduced by \cite{lakshminarayanan2017simple}, where an ensemble of deep neural networks is trained by randomly initializing a number of different models. In \cite{fort2019deep}, the authors demonstrate the stability and diversity of randomly initialized models. In \cite{wang2016ensembles}, the authors propose an (non-distributed) ensemble of GANs with different initializations to learn a distribution over images. MAD-GAN \cite{ghosh2018multi} is another solution that tackles mode-collapse, where several generators are trained together with a multi-class discriminator. 

Inspired by these previous works, in this paper we propose to build an \textbf{ensemble of fine-tuned federated generators}, illustrated in figure \ref{fig:teaser}. We emphasize that our goal is to learn to generate realistic data at the \textit{central server}, and not at the clients. Thus, our method is communication efficient and does not require more computational power from the clients, as the full ensemble is never sent out to the clients. Our proposed solution is robust with respect to mode collapse, client drift and heterogeneous data. In the work of FedGAN \cite{rasouli2020fedgan}, the authors only considered $5$ clients in their experiments. In this work, we scale this up and show that our solution works for 100 participating clients.

\section{Method}
In GAN training, two neural networks compete in a two-player minimax game. The goal of the generator $G(z;\theta_g)$ is to learn a distribution $p_g(x)$ over data $x$, by mapping input noise $z\sim p_z(z)$ to real samples $x$. Meanwhile, the discriminator $D(x;\theta_D)$ is trained to discriminate between real samples $x$ and generated samples $G(z)$. The optimization problem can thus be formulated as
\begin{equation}
    \min_G \max_D \mathbb{E}_{x\sim p_\text{data}(x)} [\log D(x)]  + \mathbb{E}_{z\sim p_z(z)}[\log (1-D(G(z)))].
\label{eq:gan}
\end{equation}

In federated learning, the data on the clients is considered sensitive and cannot be shared. Instead, a central server orchestrates the learning among clients by sending out a copy of a global model that is trained on each client's data. The new, updated models are then sent back to the central server where they are aggregated into a new global model. This procedure is repeated for a number of communication rounds until some stopping criteria is met. Let $K$ be the number of clients, $w$ the global model parameters, $x$ a data sample, $p_k$ the local data distribution for client $k$ and $\ell_k$ the loss function for client $k$. The optimization problem can then be described as

\begin{equation}
    \min_{w\in\mathbb{R}^d} \frac{1}{K} \sum_{k=1}^K \mathbb{E}_{x\sim p_k} \left[ \ell_k (w ; \; x) \right].
\label{eq:fedavg}
\end{equation}

As a baseline, we compare our method to FedGAN \cite{rasouli2020fedgan}. For FedGAN, the generator and discriminator are locally trained for a number of local epochs, optimizing equation \eqref{eq:gan}. Then, equation \eqref{eq:fedavg} is optimized using \textsc{FedAvg}, both for $G$ and $D$. That is, every local discriminator and generator is trained using SGD on the local datasets, and then every communication round the parameters are averaged to produce global models.

\subsection{EFFGAN: Ensembles of fine-tuned federated GANs}
In this work, we instead propose to build an ensemble of client generators $G_m$ for $m=1,2,\dots,M$ in order to learn the data distribution over all clients, where $M\leq K$. The ensemble thus consists of $M$ locally fine-tuned local models of the last achieved global generator using \textsc{FedAvg}. More specifically, after \textsc{FedAvg} has finished, fine-tuning on the local clients is done. Then, the $M$ number of fine-tuned generators are sent to the central server, and the data generation process is performed using all $M$ generators. Since the generation of the data happens at the central server, no extra computation is needed at the clients. An illustration of EFFGAN is shown in figure \ref{fig:teaser}. In summary our proposed method works as follows.

\begin{enumerate}
    \item Train a global generator using \textsc{FedAvg} (as in FedGAN).
    \item Fine-tune the global model achieved from 1) on $M$ randomly selected clients for $E$ number of epochs and send them back to the central server.
    \item At the server, sample a fine-tuned generator $G_m$: $m\sim\mathcal{U}(0,M)$.
    \item Sample a data point $x$ from generator $G_m$: $x\sim p^m_g(x)$.
    \item Repeat 3-4 until sufficient data has been generated.
\end{enumerate}
In our experimental setup, we will investigate how the number of fine-tuned generators $M$ affect the performance of EFFGAN, see figure \ref{fig:n_finetuned}.    

\subsection{Note on privacy} As for most federated learning methods, our proposed EFFGAN requires model information to be sent from clients to the central server during training. This is a potential privacy risk, as recent work has shown that it is possible to reconstruct training data using gradients \cite{geiping2020inverting}. Meanwhile, EFFGAN can be combined with existing works such as differential privacy \cite{geyer2017differentially}, in order to achieve stronger privacy for the clients. This is an interesting field of research but out of scope of this paper, and we leave further studies of this topic for future work.

\section{Experimental setup} 
All experiments were carried out using the DCGAN architecture \cite{radford2015unsupervised} trained on the FashionMNIST dataset \cite{xiao2017fashion} and the MNIST dataset \cite{deng2012mnist}. Meanwhile, we believe our framework is compatible with any type of GAN. DCGAN is a GAN architecture that consists of convolutional layers in the generator and convolutional-transpose layers in the discriminator. The input to the generator is a latent vector that is sampled from the standard normal distribution $z\sim\mathcal{N}(0,1)$. The DCGAN architecture was chosen for its simplicity and because it is used widely in the literature. While there today exist GANs that perform better for more complex datasets, DCGAN has sufficient capacity for the purposes of this work.

FashionMNIST contains 70,000 28x28 gray-scale images of Zalando clothing in 10 classes. It is split into 60,000 training images and 10 000 test images. MNIST contains 70,000 28x28 gray-scale images of handwritten digits, and it is also split into  60,000 training samples and 10,000 test samples.

In experimental setup of this paper, we study three main problems: heterogeneous data, client drift, and how the number of fine-tuned generators affect the ensemble performance. In order to study non-iid data, each client is assigned the same number of data points from $n$ distinct classes out of ten existing ones in the dataset. We perform experiments for varying $n$, comparing our method to the baseline. We also perform experiments varying the local number of epochs $E$ to study the effect of client drift in the non-iid setting of $n=2$. Lastly, we run experiments to investigate how sensitive EFFGAN is to the number of fine-tuned client generators $M$ that make up the ensemble.

In the beginning of each communication round, each client receives a copy of the generator ($G$) and the discriminator ($D$). Both $G$ and $D$ are then trained locally for $E$ local epochs, after which all local networks are aggregated using \textsc{FedAvg}. The models are trained using the Adam optimizer \cite{kingma2014adam}, with a learning rate of $\eta = 0.001$ and $\beta_1 = 0.0$, $\beta_2 = 0.9$. For the previous work of FedGAN \cite{rasouli2020fedgan}, the authors considered $K=5$ clients in their experiments. We scale up the experiments to $K=100$ clients and distribute $600$ samples for each client. A common way to increase efficiency in federated learning is to randomly sample a fraction $c$ of all clients in each communication round.
However, using too few clients poses the risk of loss of performance. We thus perform experiments investigating how this fraction affects the performance of our method and compare it with the baseline.

\textbf{Evaluation.} We use the Fréchet Inception Distance (FID) \cite{heusel2017gans, Seitzer2020FID} as our measure of performance, which compares the distribution of generated samples with the distribution of real ones that were used to train $G$. For FedGAN, the FID is computed after aggregation of the local models each communication round. The performance of EFFGAN is measured after the local fine-tuning, and the FID is based on data generated from the $M$ number of client generators.

\textbf{Baselines}. In our experiments, we compare EFFGAN to two main baselines. The first baseline is the aforementioned FedGAN \cite{rasouli2020fedgan}. Since EFFGAN consists of locally fine-tuned generators, the second baseline we compare with is an ensemble of locally trained generators, which did not participate in federated learning. We can thus study to what extent \textsc{FedAvg} facilitates the training of the fine-tuned generators.

\begin{figure*}
    \centering
    \begin{subfigure}[t]{0.48\textwidth}
    \includegraphics[width=\columnwidth]{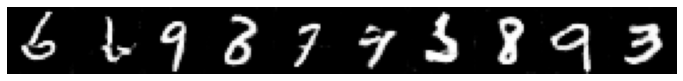}
    \includegraphics[width=\columnwidth]{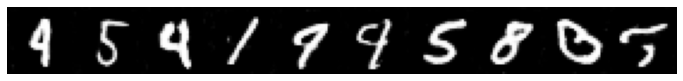}
    \includegraphics[width=\columnwidth]{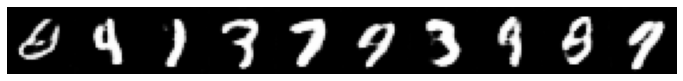}
    \caption{FedGAN}
    \label{fedgan_mnist}
    \end{subfigure}
    \begin{subfigure}[t]{0.48\textwidth}
    \includegraphics[width=\columnwidth]{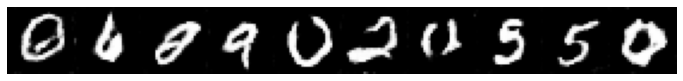}
    \includegraphics[width=\columnwidth]{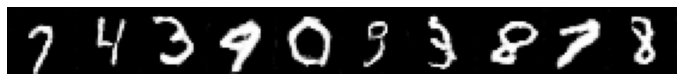}
    \includegraphics[width=\columnwidth]{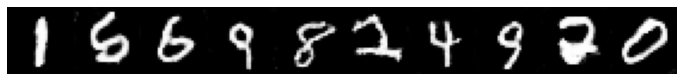}
    \caption{EFFGAN}
    \label{effgan_mnist}
    \end{subfigure}
    \caption{Images generated by (a) FedGAN \cite{rasouli2020fedgan} and (b) our EFFGAN trained on the MNIST dataset. The rows correspond to models trained with local epochs $E = 1, 5, 10$ (top to bottom).}
    \label{fig:glob_ims_mnist}
\end{figure*}

\begin{figure*}
    \centering
    \begin{subfigure}[t]{0.48\textwidth}{
    \includegraphics[width=\columnwidth]{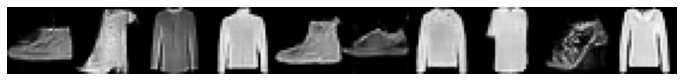}
    \includegraphics[width=\columnwidth]{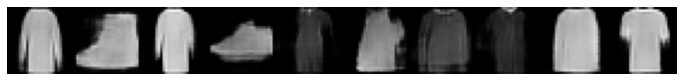}
    \includegraphics[width=\columnwidth]{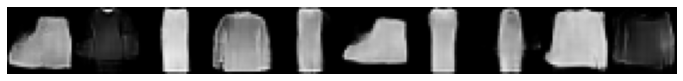}
    \includegraphics[width=\columnwidth]{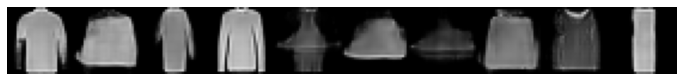}
    \includegraphics[width=\columnwidth]{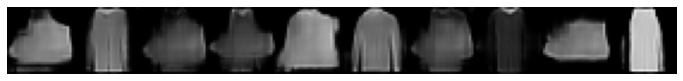}
    }
    \caption{FedGAN}
    \end{subfigure}
    \begin{subfigure}[t]{0.48\textwidth}{
    \includegraphics[width=\columnwidth]{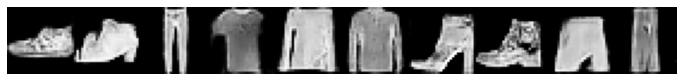}
    \includegraphics[width=\columnwidth]{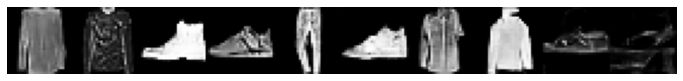}
    \includegraphics[width=\columnwidth]{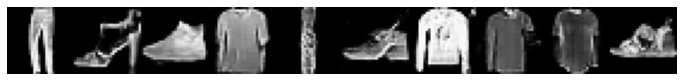}
    \includegraphics[width=\columnwidth]{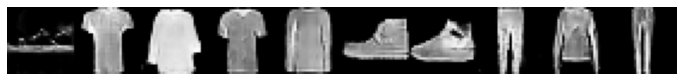}
    \includegraphics[width=\columnwidth]{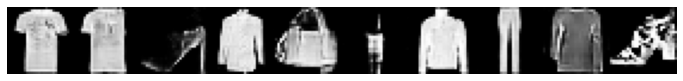}}
    \caption{EFFGAN}
    \end{subfigure}
    \caption{Images generated by (a) FedGAN \cite{rasouli2020fedgan} and (b) our EFFGAN trained on the FashionMNIST dataset. The rows corresponds to models trained with local epochs $E = 1, 5, 10, 20, 50$ (top to bottom).}
    \label{fig:glob_ims}
\end{figure*}

\section{Results} 
In this section we present the results of our work according to the experimental setup.

\subsection{Qualitative results}

Figure~\ref{fig:glob_ims_mnist} shows images generated by FedGAN and EFFGAN side by side, both approaches trained on MNIST for a varying number of local epochs. For both models, we save the best performing model with respect to FID over all communication rounds, and use that as our final model to generate data. We can see that both solutions manage to solve the task well, and generate plausible handwritten digits up to $E=5$ local epochs. For FedGAN, the generated images for $E=10$ are somewhat unclear and smudged, indicating problems of client drift.

Figure~\ref{fig:glob_ims} similarly shows images generated by FedGAN and EFFGAN side by side, using FashionMNIST for training. Here, the difference between the two methods becomes more clear. While FedGAN generates images that are blurred and grey with limited variability as the number of local epochs increases, EFFGAN consistently outputs images that are clear and with good variability. We note that EFFGAN is more robust to a larger number of local epochs, generating more realistic images up to $E=50$. This makes EFFGAN more communication efficient than FedGAN, as it can generate the same (or better) images with less communication.

\subsection{Non-iid data}
\begin{figure}[h]
    \centering
    \includegraphics[width=\columnwidth]{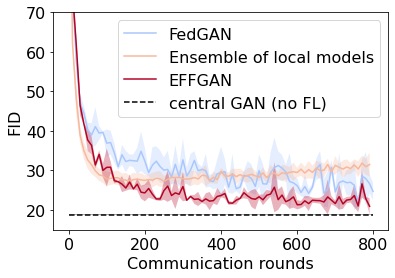}
    \caption{FID scores of different federated GAN architectures during training.}
    \label{fig:first}
\end{figure}

Figure \ref{fig:first} shows a comparison of FID scores on FashionMNIST for the FedGAN baseline (blue), our proposed ensemble of fine-tuned client models: EFFGAN (red), and an ensemble of client models trained locally without any federated learning (orange). All clients only have $n=2$ distinct classes in their datasets, making the problem non-iid. These are also benchmarked against the lowest FID reached by a centrally trained model, i.e. a model with access to the entire dataset. In this experiment, EFFGAN is an ensemble of 20 randomly chosen client generators. The results demonstrate that by using an ensemble of fine-tuned generators to generate data we reach a better FID as compared to both FedGAN and an ensemble of locally trained generators, almost reaching the performance of a centrally trained GAN. Note that a centrally trained GAN has access to all data and does not solve the distributed learning problem. These results suggest that using a generator trained with \textsc{FedAvg} serves as a good initialization for learning the local client data distributions since EFFGAN outperforms an ensemble of local models trained without communication.

\begin{figure*}
    \centering
    \includegraphics[width=\textwidth]{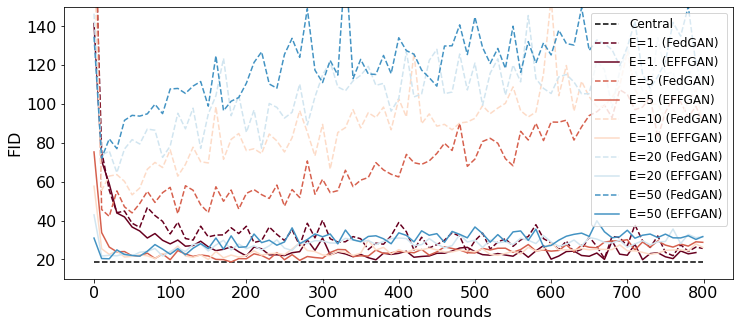}
    \caption{FID for EFFGAN (solid) and FedGAN (dashed) for different number of local epochs ($E$), trained on the FashionMNIST dataset. Here it is clearly shown that FedGAN experiences client drift as $E$ increases, whereas EFFGAN is mitigates this problem.}
    \label{fig:local_ep}
\end{figure*}

\begin{figure}[h]
    \centering
    \includegraphics[width=\columnwidth]{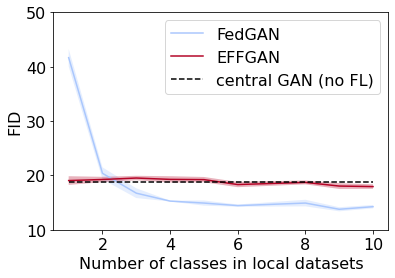}
    \caption{FID for varying heterogeneity, by varying the number of classes $n$ present in the local client datasets (higher $n \rightarrow$ more iid). Dataset: MNIST.}
    \label{fig:n_class}
\end{figure}

\begin{figure}[]
    \centering
    \includegraphics[width=\columnwidth]{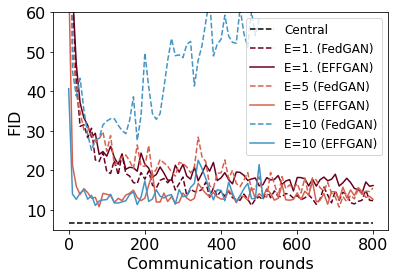}
    \caption{FID for EFFGAN (solid) and FedGAN (dashed line) for different number of local epochs ($E$), trained on the MNIST dataset.}
    \label{fig:local_ep_mnist}
\end{figure}

Figure \ref{fig:n_class} shows the advantage of using the ensemble with a varying degree of data heterogeneity, and especially so when the data distribution is highly non-iid. For this experiment, each client has at most $n$ data classes present in its local dataset, for $n = 1, 2, \dots, 10$. 
For lower values of $n$, there is a major difference between the aggregated FedGAN and our proposed ensemble EFFGAN.
As expected, this gap closes as the data distribution goes towards an iid data distribution. \textit{Most notably, EFFGAN is not affected by the data heterogeneity at all, and even performs on par with a centrally trained model when each client dataset only has samples from $n=1$ label.} This behaviour ought to be affected by the number of clients in the ensemble. As there are 10 classes in total in the dataset, and this experiment is performed with 20 clients in the ensemble, then we expect all classes to be represented by the ensemble even when there is only one class of data in the local datasets.
That is, assuming that the local generators diverge to generate data based on the local datasets, despite the occasional aggregation during training.
\begin{table}[]
\centering
\caption{The best FID score achieved by EFFGAN and FedGAN for different number of local epochs.}
\label{tab:loc_ep}
\begin{tabular}{r|r|r}
\multicolumn{1}{c}{} & \multicolumn{2}{c}{FID} \\ \hline
\multicolumn{1}{|c|}{\# local epochs $E$} & \multicolumn{1}{c|}{EFFGAN} & \multicolumn{1}{c|}{FedGAN} \\ \hline
1 & \textbf{19.8} & 20.4 \\
5 & \textbf{18.7} & 42.1  \\
10 & \textbf{19.9} & 53.3  \\
20 & \textbf{21.7} & 64.7  \\
50 & \textbf{22.0} & 72.1
\end{tabular}
\end{table}
\subsection{Client drift}
\textsc{FedAvg} is known to become detrimental when client data is distributed non-iid due to local models drifting far away from each other in parameter space when training is performed with too man local epochs. This phenomenon is known as \textit{client drift} \cite{mcmahan2017communication,karimireddy2019scaffold}.

To study the stability of our method, we investigate the performance with respect to the number of local epochs $E$ each client performs every communication round. 
In figure \ref{fig:local_ep_mnist} we observe results from this experiment for the non-iid data setting for the MNIST dataset, where each client dataset only contains $n=2$ classes. The performance of the FedGAN baseline is represented by the dashed lines, and the ensemble of fine-tuned models (EFFGAN) is represented by solid lines. Even though both FedGAN and EFFGAN seem to reach similar scores for low values of $E$, we see that an increase in $E$ leads to faster convergence for EFFGAN. Furthermore, when the number of local epochs increases to $E=10$ FedGAN starts to diverge, while EFFGAN remains stable.

In figure \ref{fig:local_ep} we see the same experiment performed for FashionMNIST. We note that for increasing $E$ here, FedGAN (dashed) performs much worse, experiencing client drift where the clients converge to different solutions in parameter space due to the heterogeneous data. This breaks \textsc{FedAvg}, resulting in performance degeneration. On the contrary, EFFGAN converges faster in terms of communication rounds towards the performance of a central GAN with an increase of $E$. \textit{More notably, it does not suffer from client drift,} which is presented as solid lines in figure \ref{fig:local_ep}.
\begin{figure}[]
    \centering
    \includegraphics[width=\columnwidth]{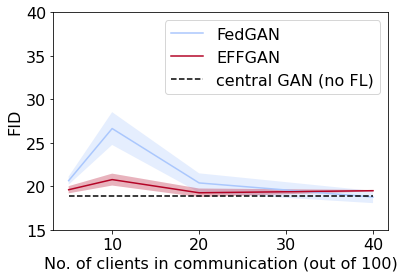}
    \caption{Varying number of clients sampled in each communication round, for the non-iid setting of $n=2$ classes in each client dataset.}
    \label{fig:comm}
\end{figure}

\begin{figure}[]
    \centering
    \includegraphics[width=\columnwidth]{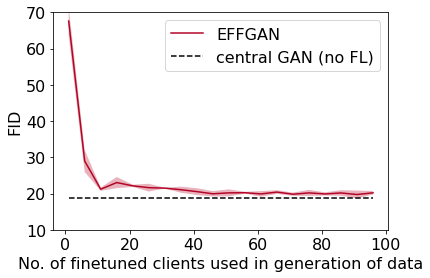}
    \caption{FID score vs the number of fine-tuned client generators that make up the ensemble in EFFGAN, in the non-iid data setting of $n=2$ classes in each client dataset.}
    \label{fig:n_finetuned}
\end{figure}
This is further presented in table \ref{tab:loc_ep}, where we report the best FID score reached by each algorithm, for a different number of local epochs. We observe here that EFFGAN reaches a better FID score as compared to FedGAN in all experiments.

In figure \ref{fig:comm} the lowest FID is recorded for a varying number of clients used in communication. Here, FedGAN tends to be lower for a larger amount of clients in communication. Meanwhile, the FID of the ensemble is stable and works well even when a small number of clients participate in each communication round. This means that less communication is needed for training EFFGAN as compared to FedGAN.

\subsection{Effect of the number of models in ensemble}
Despite having a fixed fraction of clients taking part in each communication round, we can still fine-tune all clients in the last round without reducing the efficiency of the algorithm much. 
However, an ensemble consisting of quite a few models may be unpractical and computationally unnecessary. Figure \ref{fig:n_finetuned} shows how the FID score on FashionMNIST is affected by the number of client generators that make up the ensemble. We see that the FID score decreases fast in the beginning, and there is a lot to gain from adding more generators to the ensemble. However, after the ensemble consists of $10$ generators, diminishing returns is experienced. EFFGAN can thus efficiently learn the data distribution without using a large number of client models.

\section{Discussion}
\textbf{Client drift.} A main finding in our paper is that FedGAN is sensitive to the number of local epochs used to train the generator and discriminator. This is presented in figure \ref{fig:local_ep}, where we see that FedGAN (dashed) diverges as $E$ increases, i.e. suffering from client drift. This is especially observed for the FashionMNIST dataset, which is a harder dataset distribution to learn as compared to MNIST. Meanwhile, our proposed method EFFGAN outperforms FedGAN in this regard and is especially able to perform well even when the number of local epochs is large. Whereas FedGAN suffers from client drift for large $E$, EFFGAN does not, and in our experiments, it can handle up to $50$ local epochs during training. The reason for this is that all fine-tuned members of the ensemble have learnt their own respective data distribution well, so as long as a sufficient number of clients participate in the ensemble, EFFGAN learns to model the whole data distribution. This also makes EFFGAN more communication efficient, as it does not need to communicate with the central server as often. Figure \ref{fig:n_finetuned} also shows that we do not need a high percentage of clients to build the ensemble.

For MNIST, both FedGAN and EFFGAN can successfully solve the problem and generate convincing synthetic samples. However, when increasing the number of local epochs, EFFGAN converges faster as seen in figure \ref{fig:local_ep_mnist}, and we conclude that EFFGAN requires substantially less communication. FedGAN does start to drift for $E=10$, which is avoided for EFFGAN.

\textbf{Qualitative evaluation.} In figure~\ref{fig:glob_ims_mnist} and figure~\ref{fig:glob_ims}, we can see by visual inspection that both FedGAN and EFFGAN manage to generate good samples on the simple MNIST dataset, and generate plausible handwritten digits for up to 5 local epochs. For a larger number of local epochs, the output starts to produce images that are less clear. In these figures it becomes evident that FedGAN does not handle the more complex FashionMNIST dataset well, producing increasingly grey and blurred images with low variability as the number of local epochs grows. EFFGAN on the other hand, produces convincing images for all explored numbers of local epochs, for both datasets. This indicates that the client drift that occurs in FedGAN, and which hinders the aggregation from producing a meaningful and expressive resulting model, does not occur in EFFGAN. We observe that MNIST is a simpler problem, and that FedGAN is capable of solving it with a sufficient amount of communication (local epochs at most 5). Meanwhile, EFFGAN handles both datasets quite well, regardless of the number of local epochs; thus EFFGAN requires substantially less communication overhead in comparison.
Analogously, in figure~\ref{fig:local_ep_mnist} we can see that both approaches reach a similar FID score for MNIST after training for $E=1$ and $E=5$, although as the number of local epochs increases, EFFGAN increasingly starts to converge at fewer communication rounds. For $E=10$ local epochs we start to observe client drift for FedGAN, whereas EFFGAN mitigates this.
Similar patterns can be seen in figure~\ref{fig:local_ep} for FashionMNIST. Here it becomes clear that FedGAN suffers from client drift when the number of local epochs is greater than 1. This further strengthens the conclusion that EFFGAN is less sensitive to the number of local epochs, and that it can perform well with less communication.

\textbf{Heterogeneous data.} In figure \ref{fig:n_class}, EFFGAN is compared to FedGAN in the highly non-iid setting where each client only has $n$ classes present in its dataset, for varying $n$. We see that FedGAN performs quite well for $n>2$, where data is not that heterogeneous, even beating a central GAN. This can seem counter-intuitive, but is consistent with numbers from previous work, as reported from \cite{guerraoui2020fegan}. This stems from the fact that the aggregate global generator is based on more data samples (the clients) as compared to the central model. This is a great advantage of distributed training, enhancing the learning procedure by using more diverse updates as compared to central training.

Meanwhile, EFFGAN is robust with respect to $n$, and outperforms FedGAN in the non-iid setting where $n<3$. Even in the case of $n=1$, i.e. when there is only one class out of ten in each client data set, EFFGAN manages to perform on par with a centrally trained GAN. EFFGAN is also less prone to mode collapse, i.e. only generating samples from certain modes in the data. This is due to the ensemble consisting of several fine-tuned models, each expert on their own local data distribution.

\textbf{Number of generators in ensemble.} In figure \ref{fig:n_finetuned}, we show how FID score changes as a function of the number of generators in the ensemble for the non-iid setting of $n=2$. Using very few fine-tuned generators will not be enough to capture the whole data distribution, as each generator is fine-tuned on its own local data. However, by only randomly sampling $10$ out of $100$ clients, FID is improved a lot. The ensemble is robust to increasing the number of generators, however diminishing returns of FID is experienced when adding many more.

\textbf{Transfer learning.} In figure \ref{fig:first} we note that EFFGAN outperforms both FedGAN and an ensemble of locally trained generators (with no communication). This result indicates that the generator in FedGAN achieved from \textsc{FedAvg} serves as a good initialization for transfer learning, as this is what is fine-tuned for every ensemble member in EFFGAN. This even holds true if FedGAN starts to degenerate due to client drift.


\section{Conclusions}
In this work, we have proposed a novel method for learning a data distribution over decentralized client data using an ensemble of fine-tuned federated GANs which we name EFFGAN. Our main results show that using standard \textsc{FedAvg} for training GANs works when data is iid, but suffers from client drift when data is non-iid. Our EFFGAN mitigates this problem by generating data from an ensemble of locally fine-tuned generators. First, we show in our experiments that client drift can be avoided in this way, which enables training for a large number of local epochs. This makes the proposed method communication-efficient, as we do not need to communicate with a central server often. Secondly, as long as we have a sufficient number of client generators as a part of the ensemble we are able to learn the full data distribution and achieve a FID score close to a centrally trained model even in a pathological non-iid data setting. Lastly, although FedGAN diverges in the non-iid setting when trained for several local epochs, we conclude that \textsc{FedAvg} has good transfer learning capabilities as this enables the local generators to fine-tune and learn the local data distributions better as compared to random initialization and no communication.

In this work, we experimented with non-iid data in the form of label distribution skew. For future work, it would be interesting to research other types of heterogeneity as well, such as covariate shift and concept shift. Studies investigating how the number of data samples per client dataset effect the generated images also serve as interesting future experiments.

\section*{Acknowledgment}

\bibliographystyle{IEEEtran}
\bibliography{sample-base}

\begin{thebibliography}{10}
\providecommand{\url}[1]{#1}
\csname url@samestyle\endcsname
\providecommand{\newblock}{\relax}
\providecommand{\bibinfo}[2]{#2}
\providecommand{\BIBentrySTDinterwordspacing}{\spaceskip=0pt\relax}
\providecommand{\BIBentryALTinterwordstretchfactor}{4}
\providecommand{\BIBentryALTinterwordspacing}{\spaceskip=\fontdimen2\font plus
\BIBentryALTinterwordstretchfactor\fontdimen3\font minus
  \fontdimen4\font\relax}
\providecommand{\BIBforeignlanguage}[2]{{%
\expandafter\ifx\csname l@#1\endcsname\relax
\typeout{** WARNING: IEEEtran.bst: No hyphenation pattern has been}%
\typeout{** loaded for the language `#1'. Using the pattern for}%
\typeout{** the default language instead.}%
\else
\language=\csname l@#1\endcsname
\fi
#2}}
\providecommand{\BIBdecl}{\relax}
\BIBdecl

\bibitem{mcmahan2017communication}
B.~McMahan, E.~Moore, D.~Ramage, S.~Hampson, and B.~A. y~Arcas,
  ``Communication-efficient learning of deep networks from decentralized
  data,'' in \emph{Artificial intelligence and statistics}.\hskip 1em plus
  0.5em minus 0.4em\relax PMLR, 2017, pp. 1273--1282.

\bibitem{kairouz2019advances}
P.~Kairouz, H.~B. McMahan, B.~Avent, A.~Bellet, M.~Bennis, A.~N. Bhagoji,
  K.~Bonawitz, Z.~Charles, G.~Cormode, R.~Cummings \emph{et~al.}, ``Advances
  and open problems in federated learning,'' \emph{arXiv preprint
  arXiv:1912.04977}, 2019.

\bibitem{goodfellow2014generative}
I.~Goodfellow, J.~Pouget-Abadie, M.~Mirza, B.~Xu, D.~Warde-Farley, S.~Ozair,
  A.~Courville, and Y.~Bengio, ``Generative adversarial nets,'' \emph{Advances
  in neural information processing systems}, vol.~27, 2014.

\bibitem{rasouli2020fedgan}
M.~Rasouli, T.~Sun, and R.~Rajagopal, ``Fedgan: Federated generative
  adversarial networks for distributed data,'' \emph{arXiv preprint
  arXiv:2006.07228}, 2020.

\bibitem{hardy2019md}
C.~Hardy, E.~Le~Merrer, and B.~Sericola, ``Md-gan: Multi-discriminator
  generative adversarial networks for distributed datasets,'' in \emph{2019
  IEEE international parallel and distributed processing symposium
  (IPDPS)}.\hskip 1em plus 0.5em minus 0.4em\relax IEEE, 2019, pp. 866--877.

\bibitem{yonetani2019decentralized}
R.~Yonetani, T.~Takahashi, A.~Hashimoto, and Y.~Ushiku, ``Decentralized
  learning of generative adversarial networks from non-iid data,'' \emph{arXiv
  preprint arXiv:1905.09684}, 2019.

\bibitem{augenstein2019generative}
S.~Augenstein, H.~B. McMahan, D.~Ramage, S.~Ramaswamy, P.~Kairouz, M.~Chen,
  R.~Mathews \emph{et~al.}, ``Generative models for effective ml on private,
  decentralized datasets,'' \emph{arXiv preprint arXiv:1911.06679}, 2019.

\bibitem{guerraoui2020fegan}
R.~Guerraoui, A.~Guirguis, A.-M. Kermarrec, and E.~L. Merrer, ``Fegan: Scaling
  distributed gans,'' in \emph{Proceedings of the 21st International Middleware
  Conference}, 2020, pp. 193--206.

\bibitem{hardy2018gossiping}
C.~Hardy, E.~Le~Merrer, and B.~Sericola, ``Gossiping gans: Position paper,'' in
  \emph{Proceedings of the Second Workshop on Distributed Infrastructures for
  Deep Learning}, 2018, pp. 25--28.

\bibitem{lakshminarayanan2017simple}
B.~Lakshminarayanan, A.~Pritzel, and C.~Blundell, ``Simple and scalable
  predictive uncertainty estimation using deep ensembles,'' \emph{Advances in
  neural information processing systems}, vol.~30, 2017.

\bibitem{fort2019deep}
S.~Fort, H.~Hu, and B.~Lakshminarayanan, ``Deep ensembles: A loss landscape
  perspective,'' \emph{arXiv preprint arXiv:1912.02757}, 2019.

\bibitem{wang2016ensembles}
Y.~Wang, L.~Zhang, and J.~Van De~Weijer, ``Ensembles of generative adversarial
  networks,'' \emph{arXiv preprint arXiv:1612.00991}, 2016.

\bibitem{ghosh2018multi}
A.~Ghosh, V.~Kulharia, V.~P. Namboodiri, P.~H. Torr, and P.~K. Dokania,
  ``Multi-agent diverse generative adversarial networks,'' in \emph{Proceedings
  of the IEEE conference on computer vision and pattern recognition}, 2018, pp.
  8513--8521.

\bibitem{geiping2020inverting}
J.~Geiping, H.~Bauermeister, H.~Dr{\"o}ge, and M.~Moeller, ``Inverting
  gradients-how easy is it to break privacy in federated learning?''
  \emph{Advances in Neural Information Processing Systems}, vol.~33, pp.
  16\,937--16\,947, 2020.

\bibitem{geyer2017differentially}
R.~C. Geyer, T.~Klein, and M.~Nabi, ``Differentially private federated
  learning: A client level perspective,'' \emph{arXiv preprint
  arXiv:1712.07557}, 2017.

\bibitem{radford2015unsupervised}
A.~Radford, L.~Metz, and S.~Chintala, ``Unsupervised representation learning
  with deep convolutional generative adversarial networks,'' \emph{arXiv
  preprint arXiv:1511.06434}, 2015.

\bibitem{xiao2017fashion}
H.~Xiao, K.~Rasul, and R.~Vollgraf, ``Fashion-mnist: a novel image dataset for
  benchmarking machine learning algorithms,'' \emph{arXiv preprint
  arXiv:1708.07747}, 2017.

\bibitem{deng2012mnist}
L.~Deng, ``The mnist database of handwritten digit images for machine learning
  research,'' \emph{IEEE Signal Processing Magazine}, vol.~29, no.~6, pp.
  141--142, 2012.

\bibitem{kingma2014adam}
D.~P. Kingma and J.~Ba, ``Adam: A method for stochastic optimization,''
  \emph{arXiv preprint arXiv:1412.6980}, 2014.

\bibitem{heusel2017gans}
M.~Heusel, H.~Ramsauer, T.~Unterthiner, B.~Nessler, and S.~Hochreiter, ``Gans
  trained by a two time-scale update rule converge to a local nash
  equilibrium,'' \emph{Advances in neural information processing systems},
  vol.~30, 2017.

\bibitem{Seitzer2020FID}
M.~Seitzer, ``{pytorch-fid: FID Score for PyTorch},''
  \url{https://github.com/mseitzer/pytorch-fid}, August 2020, version 0.2.1.

\bibitem{karimireddy2019scaffold}
S.~P. Karimireddy, S.~Kale, M.~Mohri, S.~J. Reddi, S.~U. Stich, and A.~T.
  Suresh, ``Scaffold: Stochastic controlled averaging for on-device federated
  learning.'' 2019.

\end{thebibliography}
\end{document}